\documentclass{article}

\usepackage{authblk}
\usepackage{fullpage}
\usepackage{xcolor}

\usepackage{microtype}
\usepackage{graphicx}
\usepackage{subfig}
\usepackage{tabu}
\usepackage{booktabs} % for professional tables
\usepackage{enumitem}
\usepackage{amsmath, amsthm}
\usepackage[numbers,sort&compress,square]{natbib}
\usepackage{pifont}
\newcommand{\cmark}{\text{\ding{51}}}%
\newcommand{\xmark}{\text{\ding{55}}}%
\usepackage{algorithm}
\usepackage{algpseudocode}

\usepackage{hyperref}
 \usepackage{comment}
 \usepackage[utf8]{inputenc}
\usepackage[english]{babel}
\makeatletter
\let\Changes@Markup@Deleted\@gobble
\newcommand{\subalign}[1]{%
  \vcenter{%
    \Let@ \restore@math@cr \default@tag
    \baselineskip\fontdimen10 \scriptfont\tw@
    \advance\baselineskip\fontdimen12 \scriptfont\tw@
    \lineskip\thr@@\fontdimen8 \scriptfont\thr@@
    \lineskiplimit\lineskip
    \ialign{\hfil$\m@th\scriptstyle##$&$\m@th\scriptstyle{}##$\hfil\crcr
      #1\crcr
    }%
  }%
}

\usepackage{Definitions}

\newcommand{\xhdr}[1]{\vspace{1mm} \noindent{{\bf #1.}}}

\graphicspath{{./FIG/}}

\begin{document}

\title{Large-scale randomized experiment reveals machine learning helps people learn and remember more effectively}

% proposed order, equal contribution.
\author[1]{Utkarsh Upadhyay$^{*}$}
\author[2]{Graham Lancashire}
\author[2]{Christoph Moser}
\author[1]{Manuel Gomez-Rodriguez}
\affil[1]{Max Planck Institute for Software Systems}
\affil[2]{Swift Management AG}

\date{}

\clubpenalty=10000
\widowpenalty = 10000

\maketitle

\newcommand\UnnumberedFootnote[1]{%
  \begingroup
  \hypersetup{hidelinks}
  \makeatletter%%
\newcommand\@mymakefnmark{\normalfont\@thefnmark\hfill}
\renewcommand\@makefntext[1]{%
%    \parindent 1em%
%    \noindent
    \hb@xt@1.8em{\hss\@mymakefnmark}#1}
\makeatother
  \renewcommand\thefootnote{}\footnote{#1}%
   \addtocounter{footnote}{-1}%
  \endgroup
}

\UnnumberedFootnote{\scriptsize $^{*}$Utkarsh Upadhyay'{}s current affiliation is Reasonal, Inc.}

\newcommand{\red}[1]{{\color{red}#1}}

\vspace{-7mm}

\begin{abstract}
Machine learning has typically focused on developing models and algorithms that would ultimately replace humans at tasks where intelligence is required. In this work, rather than replacing humans, we
focus on unveiling the potential of machine learning to improve how people learn and remember factual material. To this end, we perform a large-scale randomized controlled trial with thousands of learners
from a popular learning app in the area of mobility. After controlling for the length and frequency of study, we find that learners whose study sessions are optimized using machine learning remember the
content over $\sim$67\% longer than those whose study sessions are generated using two alternative heuristics. Our randomized controlled trial also reveals that the learners whose study sessions are optimized
using machine learning are $\sim$50\% more likely to return to the app within 4--7 days.

\end{abstract}

\section{Introduction} \label{sec:introduction}
The greater degree of control and personalization offered by learning apps and online platforms promise to facilitate the design and implementation of automated, data-driven teaching
policies that adapt to each learner'{}s knowledge over time, improving upon the traditional one-size-fits-all human instruction. However, to fulfill this promise, it is necessary to develop
adaptive data-driven models of the learners, which accurately quantify their knowledge, and efficient methods to find teaching policies that are provably optimal under the learners'{}{}
models~\cite{mozer2019artificial, sense2019perspectives}.

In this context, research in the (theoretical) computer science literature has been typically focused on finding teaching policies that enjoy optimality guarantees under simplified mathematical
models of the learner'{}s knowledge~\cite{lewis2014combinatorial, nishimura2018critically, novikoff2012education, reddy2016unbounded, hunziker2019teaching}. In contrast, research in
cognitive sciences has focused on measuring the effectivity of a variety of heuristic teaching policies informed by psychologically valid models of the learner'{}s knowledge using (small)
randomized control trials~\cite{pavlik2008using, metzler2009does, lindsey2014improving, settles2016trainable}. Only very recently, Tabibian et al.~\cite{tabibian2019enhancing} has introduced
a machine learning modeling framework that bridges the gap between both lines of research---their framework can be used to determine the optimal rate of study a learner should follow under
a model of the learner'{}s memory state that is informed by real human memory data. However, in the evaluation of their framework, the authors resort to a natural experiment using data from
a popular language-learning online platform rather than a randomized control trial, the gold standard in the cognitive sciences literature. As a result, it has been argued that, in an interventional
setting, an actual learner following the rate of study may fail to achieve optimal performance~\cite{mozer2019artificial}.

In this paper, we build upon the modeling framework of Tabibian et al.~\cite{tabibian2019enhancing} and design \textsc{Select}, a simple, efficient and adaptive machine learning algorithm with theoretical
guarantees to determine which questions to include in a learner'{}s sessions of study over time, rather than optimizing the rate of study as in Tabibian et al., which is typically chosen by the learner.
Then, we perform a large-scale randomized controlled trial involving 50,700 learners from a popular learning app in the area of mobility to quantify to evaluate what extent our algorithm can help
people learn and remember more effectively. By the end of the randomized controlled trial, we recorded more than $\sim$16.75 million answers to $\sim$1,900 questions in $\sim$628,000 study
sessions. After controlling for how long learners study a question and how many times they review the question, we find that learners whose study sessions are optimized using our machine learning
algorithm remember the content over 92\% longer than those whose study sessions are generated at random and 67\% longer than those whose study sessions are generated using smarter heuristic,
which has been already used in production before our randomized controlled trial. Our randomized controlled trial also reveals that the learners whose study sessions are optimized using our
algorithm are more engaged. More specifically, they are $\sim$50\% more likely to return to the app within 4-7 days. To facilitate research at the intersection of cognitive science and machine
learning, we are releasing open-source implementation of our algorithm and all the data gathered during our randomized control trial at \url{https://github.com/Networks-Learning/spaced-selection/}.

\section{Methods} \label{sec:methods}
Given a set of questions I whose answers a learner wants to learn, we represent each study session as a triplet $e := (t,\Scal,r_\Scal)$, where $\Scal \subseteq \Ical$ is the set of questions that
the learner reviewed at time $t$ and $r_\Scal$ is a vector in which each entry corresponds to a question in the set $\Scal$ and indicates whether the learner recalled ($r=1$) or forgot ($r=0$)
the answer to the question. Here, note that in the learning app that we used in our randomized experiment, the learner is tested in each study session, similar to most spaced repetition software
and online platforms such as Mnemosyne, Synap, and Duolingo, and the seminal work of Reidiger and Karpicke~\cite{roediger2006test}.

Given the above representation, we keep track of the study times using a counting process $N(t)$, which counts the number of study sessions up to time $t$. Following the literature on temporal
point processes~\cite{aalen2008survival}, we characterize this counting process using its corresponding intensity $u(t)$, \ie, $E[dN(t)]= u(t)dt$, and think of the set of questions $\Scal$ and vector
$r_\Scal$ as its binary marks.
Moreover, we utilize two well-known memory models from the psychology literature, the exponential and the power-law forgetting curve models with binary recalls~\cite{wixted2007wickelgren, averell2011form, ebbinghaus2013memory, loftus1985evaluating}, to estimate the probability $m_i(t)$ that a learner recalls (forgets) the answer to a question $i$ at time $t$.
Under both the exponential and the power-law models, the recall probability depends on the time since the last review $\Delta_i(t)$ and the forgetting rate $n_i (t) \in \RR^{+}$, which may depend on many
factors, \eg, number of previous (un)successful recalls of the answer to the question.
To estimate the value of the forgetting rate $n_i(t)$, we use (a variant of) half-life regression~\cite{settles2016trainable}. Half-life regression implicitly assumes that: (i) each question has an initial forgetting
rate $n_i (0)$, which captures the difficulty of the question; (ii) a successful recall of the answer to a question $i$ at time $t'$ during a review change the forgetting rate by $(1-\alpha_i)$, \ie, $n_i (t)= (1-\alpha_i ) n_i (t'),\, 0 \leq \alpha_i \leq 1$; and, (iii) an unsuccessful recall changes the forgetting rate by $(1+\beta_i)$, \ie, $n_i (t)= (1+\beta_i ) n_i (t'),\, \beta_i \geq 0$.

Finally, given a set of questions $\Ical$, we cast the optimization of the study sessions as the search for the optimal selection probabilities $p_i (t) := \PP[i \in \Scal]$ for each question $i \in \Ical$ that minimize
the expected value of a particular (convex) loss function $l(\mb(t),\nbb(t), \mathbf{\Delta}(t),\pb(t))$ of the recall probability of the answers to the questions $\mb(t) = [ m_i(t) ]_{i \in \Ical}$, the forgetting rates
$\nbb(t) = [n_i(t)]_{i \in \Ical}$, the times since their last review $\mathbf{\Delta}(t) = [\Delta_i(t)]_{i \in \Ical}$, and the selection probabilities $\pb(t)= [p_i(t)]_{i \in \Ical}$ over a time window $(t_0,t_f]$, \ie,
\begin{equation} \label{eq:optimization}
\text{minimize}_{p(t_0,t_f]} \quad \EE \left[ \phi(\mb(t), \nbb(t), \Delta(t)) + \int_{t_0}^{t_f} l(\mb(\tau),\nbb(\tau), \mathbf{\Delta}(\tau),\pb(\tau)) \, d\tau \right]
\end{equation}
where $p(t_0,t_f]$ denotes the selection probabilities from $t_0$ to $t_f$, the expectation is taken over all possible realizations of the selection probabilities, the counting process $N(t)$ and the recalls of the answers
to the questions, the loss function is nonincreasing (nondecreasing) with respect to the recall probabilities and the times since their last review (forgetting rates and selection probabilities) so that it rewards long-lasting
learning while limiting the number of reviews, and $\phi(\mb(t), \nbb(t), \Delta(t))$ is an arbitrary penalty function. Here, note that the rate of study session $u(t)$ is unknown and is not under our control, \ie, the learner
chooses when to study.

To solve the optimization problem defined by Eq.~\ref{eq:optimization}, we proceed similarly as in Tabibian et al.~\cite{tabibian2019enhancing} and resort to the theory of stochastic optimal control of jump SDEs~\cite{hanson2007applied}. However, in contrast with Tabibian et al., rather than optimizing the rate of study, we optimize the selection probability of each question in each study session. More specifically,
if we penalize quadratically the probability of unsuccessful recall of the answer to a question upon review and the probability of studying the question, \ie,
\begin{equation}
l(\mb(t),\nbb(t), \mathbf{\Delta}(t),\pb(t)) = \sum_{i \in \Ical} \frac{1}{2} (1-m_i (t))^2 u(t)+ \frac{1}{2} q \, p_i^2 (t)  u(t),
\end{equation}
where $q \geq 1$ is a given parameter, which trades off recall probability upon review and the size of the study sessions---the higher its value, the shorter the study sessions. Then, we can show that, for each
question $i \in \Scal$, the optimal selection probability is given by (refer to Appendix~\ref{app:theory} for more details)\
\begin{equation} \label{eq:optimal}
p_i^* (t)=  \frac{1}{\sqrt{q}} (1-m_i (t))
\end{equation}
Finally, since the optimal selection probability depends only on the recall probability, which is estimated either the exponential or the power-law forgetting curve model, we can implement a very efficient procedure
to construct study sessions, which we name \textsc{Select}. Algorithm~\ref{alg:select} provides a pseudocode implementation of \textsc{Select}.
\begin{algorithm}[t]                    % enter the algorithm environment
  \caption{\textsc{Select} Find the probability of selection of an item for study for one learner}
  \label{alg:select}
%  \setstretch{1.5}
  \begin{algorithmic}[1]
  \State \textbf{Input: } Set of items $\Jcal$, number of (un)successful recalls $n^{\cmark}_i(t)$ and $n^{\xmark}_i(t)$, initial difficulties $\{ n_i(0) \}$, last re\-view times $\{ t_i \}$ and parameters $\alpha$, $\beta$ and $q$.
  \State \textbf{Output: } Probability of selection of each item $\mathbf{p}(t)$.

  \State $\mathbf{p}(t) \leftarrow \mathbf{0}$
  \For {$i \in \Jcal$}
    \State $n_i(t) \leftarrow n_i(0) (1 - \alpha)^{n^{\cmark}_i(t)} (1 + \beta)^{n^{\xmark}_i(t)}$
    \State $m_i(t) \leftarrow \exp{ \left( -n_i(t) (t - t_i) \right) }$ \Comment Using exponential memory model
    \State $p_i(t) = \frac{1}{\sqrt{q}}(1 - m_i(t))$
  \EndFor

%  \State $\Pbar \leftarrow \sum_{i \in \Jcal} p_i

  \State \Return $\mathbf{p}(t)$

%   \State $F_0 \leftarrow $ \textbf{\textsc{Evaluate}}$(\Xcal_0)$
%
%       \For {$g \in [0, \dots, G - 1]$}
%           \State $U \leftarrow$ \textsc{SelTournamentCDC}$(\Xcal_g, F_g)$ \Comment{Select off-springs using tournaments~\cite{deb2002}}
%           \State $V \leftarrow \varnothing$
%
%           \State $Z_1 \leftarrow$ \textsc{RandomChoose}$(U, P / 2)$ \Comment Partition $U$ into 2 equal sets
%           \State $Z_2 \leftarrow Y \cap Z_1$
%
%           \For {$i \in [1, \dots, P / 2]$}
%               \State $o_1 \leftarrow $ \textsc{RandomChoose}$(Z_1, 1)$
%               \State $o_2 \leftarrow $ \textsc{RandomChoose}$(Z_2, 1)$
%
%               \State $Z_1 \leftarrow Z_1 \setminus \{ o_1 \}$
%               \State $Z_2 \leftarrow Z_2 \setminus \{ o_2 \}$
%
%               \If {\textsc{Random}$() < p_x$}
%                  \State $o_1, o_2 \leftarrow $ \textsc{Mate}$(o_1, o_2)$
%               \EndIf
%
%               \State $V \leftarrow V \cup \{ o_1, o_2 \}$
%           \EndFor
%
%           \State $\Xcal_{g + 1} \leftarrow \varnothing$
%           \For {$o \in V$}
%               \If {\textsc{Random}$() < p_m$}
%                   \State $o \leftarrow $ \textsc{Mutate}$(o)$ \Comment Random binary noise for each bit
%               \EndIf
%
%               \State $\Xcal_{g + 1} \leftarrow \Xcal_{g + 1} \cup \{ o \}$
%           \EndFor
%
%           \State $F_{g + 1} \leftarrow $ \textbf{\textsc{Evaluate}}$(\Xcal_{g + 1})$
%     \EndFor
%
%     \State \Return \textsc{Pareto}$\left( \bigcup_{g \in [0, \dots, G]} \left\{ \Xcal_g \right\} \right)$
  \end{algorithmic}
\end{algorithm}

\section{Experimental Design} \label{sec:design}
We conduct a randomized controlled trial with all learners of at least 18 years of age in Germany who signed up for iTheorie F\"{u}hrerschein Auto, a popular app to study for the written
portion of the driver'{}s permit, from December 2019 to July 2020 (refer to Appendix~\ref{app:itheorie} for additional details on the iTheorie F\"{u}hrerschein Auto).

Before the start of our randomized controlled trial, we recorded the study sessions of all the learners who used the app from February 2019 to June 2019 to estimate the parameters of the
memory models using a variant of half-life regression~\cite{settles2016trainable}, as discussed in Methods. Here, we fit a single set of parameters $\alpha$ and $\beta$ for all questions and a different
initial forgetting rate $n_i(0)$ per question (refer to Appendix~\ref{app:prediction} for a series of benchmarks and evaluations for the fitted memory models). We define a session as a continuous period
of question answering with a gap of less than 5 minutes between answers.

During the randomized controlled trial, each learner was randomly assigned to a `select' group, a `difficulty' group, or a `random' group throughout her entire usage of the app. In the select group
($n = 10{,}151$ learners), the questions of each study session are chosen according to the optimal selection probabilities $p_i^*(t)$ under the fitted exponential forgetting curve model. In the difficulty
group ($n = 34{,}029$), they are chosen in circular order proportionally to the initial difficulty $n_i(0)$, \ie, easier questions first.
In the random group ($n = 13{,}600$), they are chosen uniformly at random. By the end of the randomized controlled trial, we recorded more than $\sim$$16.75$ million answers to $\sim$$1{,}900$ questions
by $\sim$$50{,}700$ learners in $\sim$$628{,}000$ study sessions.

For consistency, we remove the data from the $6{,}774$ learners who re-installed the app during the trial period and were assigned to a different group after the re-installation (or installed the app on different devices).
Moreover, since we do not expect any algorithm, including ours, to help learners who are cramming for tests, we skip data from the $32{,}445$ learners who used the app for less than 2 days.
After these preprocessing steps, the resulting dataset contains data from $\sim$$313{,}000$ study sessions by $11{,}481$ learners. Moreover, for each group (select, difficulty or random), the dataset contains
$\sim$$894{,}000$, $\sim$$3.3$ million and $\sim$$693{,}000$ unique (learner, question) reviewing sequences due to $1{,}564$, $7{,}582$ and $2{,}335$ learners, respectively.
This indicates a reduction of about  78\% for `select', 73\% for `difficulty', and 74\% for `random'.
This discrepancy in the relative reduction can partially be explained by a corresponding slight decrease in the number of crash-free users everyday,
ie, a median decrease of 0.17\% per day between `difficulty' and `select'.
The marginally higher complexity of the \textsc{Select} algorithm initially caused over/under flow problems which could have prompted learners to stop using the app before they reached the 2 days of learning activity.

For each (learner, question) reviewing sequence, we compare learners from each of the groups using the empirical forgetting rate~\cite{tabibian2019enhancing}, defined as $\hat{n} = -(\log \hat{m}(t_n)) / (t_n-t_{n-1})$,
where $t_n-t_{n-1}$ is the last retention interval and $\hat{m}(t)=\max(\epsilon,\min(1-\epsilon,r(t)))$, with $\epsilon = 0.01$, indicates whether the learner was able to or unable to remember the answer at time $t$ (The
results presented are agnostic to the exact constant $\epsilon$ chosen here).
Moreover, for a fairer comparison across questions, we normalize each empirical forgetting rate using the average empirical initial forgetting rate of the corresponding question at the beginning of the observation
window across learners who studied the question, and control for the number of attempts learners have made at answering the question (and, hence, seen the answer) and the duration for which they have used the app.

\section{Results} \label{sec:results}
\begin{figure}[t]
  \centering
  \subfloat[T $= 3 \pm 0.8$ days]{\includegraphics[width=0.3\textwidth]{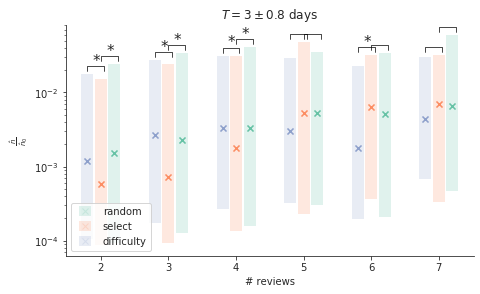}\label{fig:T-3}}\hspace{0.04\textwidth}%
  \subfloat[T $= 5 \pm 1.2$ days]{\includegraphics[width=0.3\textwidth]{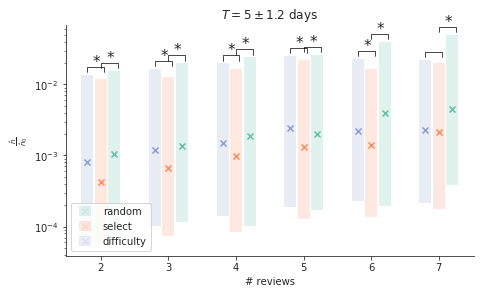}\label{fig:T-5}}\hspace{0.04\textwidth}%
  \subfloat[T $= 9 \pm 2.2$ days]{\includegraphics[width=0.3\textwidth]{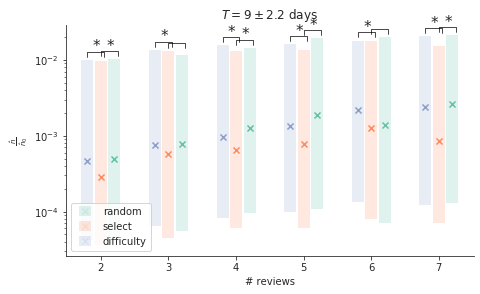}\label{fig:T-9}}
  \caption{\textbf{Normalized empirical forgetting rate} (lower is better). Each triplet of bars in the figures corresponds to (learner, question) pairs in which the learner reviewed the question the same number of times (\# reviews) for approximately the same period of time (T). Boxes indicate 25\% and 75\% quantiles and crosses indicate median values, where lower values indicate better performance. For each triple of bars, $*$ indicates a statistically significant difference (Matt-Whitney U-test; P-value = 0.05).}\label{fig:T}
\end{figure}

We first compare learners of the `select', `difficulty' and `random' groups in terms of normalized empirical forgetting rate. Figure~\ref{fig:T} summarizes the results, where each triplet of bars in the figures corresponds to (learner, question) pairs in which the learner reviewed the question the same number of times during approximately the same period of time. For example, the bar corresponding to \# reviews = 5 in Figure~\ref{fig:T-5}  contains (learner, question) pairs in which the learner reviewed the question five times during $5 \pm 1.2$ days. Moreover, note that, in the y-axis, the scale is logarithmic and the sign * indicate statistically significant differences (Matt-Whitney U-test; $p$-value $= 0.05$).

The results show that, in 83.5\% of the cases, the empirical forgetting rate for the learners in the `select' group is lower than that of the learners in the `difficulty' and `random' groups and, in 75\% of cases, the decrease is statistically significant. Moreover, the median empirical forgetting rate for learners in the `select' group is 48\% lower than that of learners in the `random' group and 40\% lower than that of learners in the `difficulty' group. In other words, the Select algorithm will help a median learner retain the answer to a median question 92\% and 67\% longer, respectively.

Next, we evaluate to what extent our algorithm can help increase (learner) engagement. To this end, we compare learners of the `select', `difficulty' and `random' groups using Firebase Analytics. Table~\ref{tab:engagement} summarizes the results, where \emph{engagement} (\emph{retention} in Firebase Analytics) indicates how likely is that a learner returns to the app within 4--7 days. We can conclude that, in terms of engagement, learners of the `select' (`difficulty') group were 50.6\% (47.6\%) more likely, in median, to return to the app within 4--7 days than learners of the `random' group. Moreover, Firebase Analytics estimates that, with probability 80\%, the \textsc{Select} algorithm is the top performer in terms of engagement.
\begin{table}
\centering
\begin{tabular}{ccc}
\toprule
\textbf{Algorithm} & \textbf{Median improvement in engagement (95\%-ile)} & \textbf{Probability to be best} \\ % & \textbf{Users} \\
\midrule
{random} & Baseline & $< 0.1\%$ \\ % & 12,029\\
{difficulty} & +46.7\% (+40.9\% to +54.7\%) & 20\% \\ % & 7,694\\
{select} & +50.6\% (+43.7\% to +57.7\%) & 80\% \\ %& 7,640
\end{tabular}
\caption{\textbf{Learner engagement for the `random', `difficulty' and `select' groups.} Engagement indicates how likely is that a learner returns to the app within 4--7 days. The 95\%-ile indicates the range in which the real value of the increase in engagement over baseline is contained with 95\% probability.}\label{tab:engagement}
\end{table}

\vspace{3mm}

\xhdr{Acknowledgements} We thank Robert West, Klein Lars Henning, Roland Aydin, and Behzad Tabibian for helpful conversations.

\bibliographystyle{unsrt}
\bibliography{refs}

\appendix

\section{Finding the optimal selection probabilities} \label{app:theory}
In this section, we derive the optimal selection probabilities $\mathbf{p}(t)$ for the power-law memory model [16, 17] following a similar
proof technique as in Tabibian et al. [12]. The derivation for the exponential memory model [14, 15] can be done similarly. 

First, we express the dynamics of the forgetting rates $n_i(t)$ using the following stochastic differential equation (SDE) with jumps: 
\begin{equation} \label{eq:forgetting-rate}
dn_i(t) = -\alpha_i n_i(t) r_i(t) dN_i(t) + \beta n_i(t)(1-r_i(t)) dN_i(t)
\end{equation}
where $E[dN_i(t)] = p_i(t) u(t)\, dt$. Now, we can use the above expression, Eq. 1 in the main paper, 
and It\"{o}'{}s calculus [26] to express the dynamics of the recall probabilities $m_i(t)$ and the times 
since the last reviews $\Delta_i(t)$ using also SDEs with jumps, \ie,
\begin{align} 
dm_i(t) &= -\frac{n_i(t)m_i(t)dt}{(1+ \Delta_i(t))} + (1-m_i(t)) dN_i(t) \label{eq:recall-probability-dm} \\
d\Delta_i(t) &= dt -\Delta_i(t) dN_i(t) \label{eq:time-review}
\end{align}
Then, given the above expressions, we can decompose the optimization problem defined by Eq. 2 in the main 
paper into $K$ independent problems, \ie,
\begin{equation} \label{eq:spaced-repetition-problem-one-item}
\underset{p_i(t_0,t_f]}{\text{minimize}} ~~\EE_{(N_i, r_i)(t_0,t_f]} \big[ \phi(m_i(t_f), n_i(t_f), \Delta_i(t_f)) +
\int_{t_0}^{t_f} \ell(m_i(\tau), n_i(\tau), \Delta_i(\tau), p_i(\tau))\, d\tau \big], % \nonumber\\
% \text{subject to} & ~~0 \leq p_i(t) \leq 1 ~ \forall t \in (t_0,t_f),
\end{equation}
which can be solved separately. 

Given a fixed question $i$, we denote $m(t) = m_i(t)$, $n(t) = n_i(t)$, $\Delta(t) = \Delta_i(t)$ and $p(t) = p_i(t)$, define the optimal 
cost-to-go function $J(m(t),n(t),\Delta(t),t)$ for the corresponding optimization problem as
\begin{equation}
J(m(t),n(t),\Delta(t),t) = \underset{p(t,t+dt]}{\text{min}} \EE[J(m(t+dt),n(t+dt),\tau(t+dt),t+dt)] + \ell(m(t),n(t),\Delta(t),p(t))dt  \label{eq:cost-to-go}
\end{equation}
and use Bellman'{}s principle of optimality to derive the corresponding HJB equation [26]. In particular, we can first 
rewrite the above equation as
\begin{equation}
0 = \underset{p(t,t+dt]}{\text{min}} \EE[dJ(m(t),n(t),\Delta(t),t)] + \ell(m(t),n(t),\Delta(t),p(t))dt, \label{eq:bellman}
\end{equation}
where $dJ(m(t),n(t),\Delta(t),t) = J(m(t+dt),n(t+dt),\Delta(t+dt),t+dt) - J(m(t),n(t),\Delta(t),t)$,
and then use the following technical Lemma, which can be proved using It\"{o}'{}s calculus [26], to differentiate $J$ with respect 
to their parameters.
\begin{lemma} \label{lem:ito}
Let $x(t)$, $y(t)$, $k(t)$ be three jump-diffusion processes defined by the following jump SDEs:
\begin{align*}
dx(t) =& f(x(t), y(t), k(t), t)dt + g(x(t),y(t),k(t),t)z(t)dN(t) + h(x(t),y(t),k(t),t)(1-z(t))dN(t)\\
dy(t) =& p(x(t),y(t),k(t),t)dt + q(x(t),y(t),k(t),t)dN(t)\\
dk(t) =& s(x(t),y(t),k(t),t)dt + v(x(t),y(t),k(t),t)dN(t)
\end{align*}
where $N(t)$ is a jump process and $z(t) \in \{0, 1\}$. If function $F(x(t), y(t), k(t), t)$ is once continuously differentiable in $x(t)$, $y(t)$, $k(t)$,
and $t$, then,
\begin{align*}
dF(x(t), y(t),k(t), t) &= (F_t + f F_x + p F_y + s F_k)(x(t), y(t), k(t), t)dt \\
&+\left[F(x+g,y+q,k+v,t)z(t) +F(x+h,y+q,k+v,t)(1-z(t)) -F(x,y,t)\right]dN(t),
\end{align*}
where for notational simplicity we dropped the arguments of the functions $f$, $g$, $h$, $p$ and $q$.
\end{lemma}
More specifically, we can write $dJ(m(t),n(t),\Delta(t),t)$ in Eq.~\ref{eq:bellman} as
\begin{align*}
dJ(m(t),n(t),\Delta(t),t) &= J_t(m(t),n(t),\Delta(t),t)- \frac{n(t)m(t)}{\Delta(t)+1}J_m(m(t),n(t),\Delta(t),t) + J_{\tau}(m,n,\tau,t) \\
&+ [J(1,(1-\alpha)n(t),0,t)r(t)  +J(1,(1+\beta)n(t),0,t)(1-r(t))-\\
&~~~~J(m(t),n(t),\Delta(t),t)]dN(t).
\end{align*}
and thus write the HJB equation as:
\begin{align}
0 &= J_t(m(t),n(t),\Delta(t),t)- \frac{n(t)m(t)}{\Delta(t)+1}J_m(m(t),n(t),\Delta(t),t) + J_{\tau}(m(t),n(t),\Delta(t),t) \nonumber \\
&+ \underset{u(t,t+dt]}{\text{min}} \big\{ \ell(m(t),n(t),u(t)) \nonumber \\
&+ \left[J(1,(1-\alpha)n(t),0,t)m(t)  +J(1,(1+\beta)n(t),0,t)(1-m(t))-J(m(t),n(t),\Delta(t),t)\right] p(t) u(t) \big\} \label{eq:hjb}
\end{align}
To solve the above differential equation, we need to define the loss $\ell$. Following the literature on stochastic
control [26], we consider the following quadratic form, which penalizes quadratically the probability of unsuccessful 
recall of the answer to the question upon review and the probability of reviewing the question:
\begin{equation} \label{eq:loss}
\ell(m(t),n(t),\Delta(t),p(t)) = \frac{1}{2} (1-m(t))^2 u(t) + \frac{1}{2} q\, p^2(t) u(t)
\end{equation}
Now, if we plug in the above loss into the HJB equation and set its derivative with respect to $p(t)$ to zero to derive the 
optimal $p^*(t)$, we obtain:
\begin{align}
p^*(t) &= \frac{1}{q} \left[ J(m(t),n(t),\Delta(t),t)-J(1,(1-\alpha)n(t),0,t)m(t) - J(1,(1+\beta) n(t),0,t)(1-m(t)) \right]_{(0, q)}, \label{eq:optimal-p}
\end{align}
where the operator $[\cdot]_{(a, b)} := \min{(\max{(a, \cdot)}, b)}$ is required to ensure that $0 \leq p(t) \leq 1$.
However, we will simplify these constraints to only positivity constraint and later verify that, by appropriately setting the tuning parameter $q$, we 
can make $p^*(t) \leq 1$ as well. Hence, if we plug in
\begin{equation}
p^*(t) = \frac{1}{q} \left[J(m(t),n(t),\Delta(t),t)-J(1,(1-\alpha)n(t),0,t)m(t)  - J(1,(1+\beta) n(t),0,t)(1-m(t)) \right]_{+}. \label{eq:optimal-p-positive}
\end{equation}
back into the HJB equation, we find that the optimal cost-to-go J needs to satisfy the following nonlinear differential equation:
\begin{align}
0 &= J_t(m(t),n(t),\Delta(t),t)- \frac{n(t)m(t)}{\Delta(t)+1}J_m(m(t),n(t),t) +  J_{\tau}(m(t),n(t),\Delta(t),t) + \frac{1}{2} (1-m(t))^2 \nonumber \\
&- \frac{u(t)}{2q} \left[ J(m(t),n(t),\Delta(t),t)-J(1,(1-\alpha)n(t),0,t)m(t)
-J(1,(1+\beta)n(t),0,t)(1-m(t))\right]_{+}^2. \nonumber
\end{align}
To solve it, we rely on the following technical Lemma:
\begin{lemma}\label{lem:hjb-proposals}
Consider the following family of losses with parameter $d > 0$,
\begin{align}
\ell_d(m(t),n(t),p(t)) &=  h_d(m(t),n(t),\Delta(t)) + g_d^2(m(t),n(t)) + \frac{1}{2} q\, p^2(t) u(t), \nonumber \\
g_d(m(t),n(t)) &= \sqrt{\frac{u(t)}{2}} \left[c_2 \frac{\log(d)}{-m(t)^2 + 2m(t) - d} - c_2\frac{\log(d)}{1-d} + c_1 m(t) \log\left( \frac{1+\beta}{1-\alpha}\right) - c_1\log(1+\beta)\right]_{}, \nonumber \\
h_d(m(t),n(t),\Delta(t)) &= - {\sqrt{q}} \frac{m(t) n(t)}{1+\Delta(t)} c_2 \frac{(-2m(t) + 2) \log(d)}{(-m(t)^2+2m(t)-d)^2}
% + \frac{\sqrt{q}\,u'(t)}{u^2(t)} \left(c_1 \log(n(t)) + c_2 \frac{\log(d)}{-m(t)^2 + 2m(t) - d}\right)
\label{eq:family}
\end{align}
where $c_1, c_2 \in \RR$ are arbitrary constants. Then, the cost-to-go $J_d(m(t),n(t), \Delta(t), t)$ that satisfies
the HJB equation, defined by Eq.~\ref{eq:hjb}, is given by:
\begin{align}
% J_d(m(t),n(t),\Delta(t),t) = \frac{\sqrt{q}}{u(t)} \left(c_1 \log(n(t)) + c_2 \frac{\log(d)}{-m(t)^2 + 2m(t) - d}\right) \label{eq:family-cost-to-go}\\
% \frac{\partial J_d(m(t),n(t),\Delta(t),t)}{\partial t} = - \frac{\sqrt{q}\,u'(t)}{u^2(t)} \left(c_1 \log(n(t)) + c_2 \frac{\log(d)}{-m(t)^2 + 2m(t) - d}\right)
J_d(m(t),n(t),\Delta(t),t) &= {\sqrt{q}} \left(c_1 \log(n(t)) + c_2 \frac{\log(d)}{-m(t)^2 + 2m(t) - d}\right) \label{eq:family-cost-to-go}\\
\implies \frac{\partial J_d(m(t),n(t),\Delta(t),t)}{\partial t} &= 0\\
\frac{\partial J_d(m(t),n(t),\Delta(t),t)}{\partial \Delta} &= 0\\
\frac{\partial J_d(m(t),n(t),\Delta(t),t)}{\partial m(t)} &= - c_2 \frac{(-2m(t) + 2) \log(d)}{(-m(t)^2+2m(t)-d)^2}
\end{align}
and the optimal intensity is given by:
\begin{equation*}
p^*(t) = q^{-1/2} \left[c_2 \frac{\log(d)}{-m(t)^2 + 2m(t) - d} - c_2\frac{\log(d)}{1-d} + c_1 m(t) \log\left( \frac{1+\beta}{1-\alpha}\right) - c_1\log(1+\beta)\right]_{+}.
\end{equation*}

\begin{proof}
Consider the family of losses defined by Eq.~\ref{eq:family} and the functional form for the cost-to-go defined by Eq.~\ref{eq:family-cost-to-go}.
Then, for any parameter value $d > 0$, the optimal intensity $p^{*}_d(t)$ is given by
\begin{align*}
p_d^*(t) &= \frac{1}{q} \left[J_d(m(t),n(t),\Delta(t),t)-J_d(1,(1-\alpha)n(t),0,t)m(t)  -J_d(1,(1+\beta) n(t),0,t)(1-m(t)) \right]_{+} \\
&= \frac{1}{\sqrt{q}} \left[ c_2 \frac{\log(d)}{-m^2 + 2m - d}  - c_2\frac{\log(d)}{1-d} + c_1 m(t) \log\left( \frac{1+\beta}{1-\alpha}\right) - c_1\log(1+\beta)\right]_{+},
\end{align*}
and the HJB equation is satisfied:
\begin{align*}
&\frac{\partial J_d(m,n,\Delta,t)}{\partial t} - \frac{mn}{1+\Delta} \frac{\partial J_d(m,n,\Delta,t)}{\partial m} + \frac{\partial J_d(m,n,\Delta,t)}{\partial \Delta} + h_d(m,n,\Delta) + g_d^2(m,n) \\
&-\frac{u}{2q} \left[J_d(m,n,\Delta,t)-J_d(1,(1-\alpha)n,0,t)m - J_d(1,(1+\beta)n,0,t)(1-m)\right]_{+}^2\\
&= \frac{mn}{1+\Delta} \underbrace{\sqrt{q} c_2 \frac{(-2m + 2) \log(d)}{(-m^2+2m-d)^2}}_{\frac{\partial J_d(m, n, \Delta, t)}{\partial m}} + h_d(m,n,\Delta) + g_d^2(m,n) \\
&\quad -\frac{u}{2}\left[ c_1 \log( n ) + c_2 \frac{\log(d)}{-m^2 + 2m - d} - m \left(c_1 \log(n(1-\alpha)) + c_2 \frac{\log(d)}{1 - d} \right) - (1-m) \left(c_1 \log(n(1+\beta))  + c_2 \frac{\log(d)}{1 - d}\right) \right]_{+}^2 \\
&= \frac{mn}{1+\Delta} \sqrt{q} c_2 \frac{(-2m + 2) \log(d)}{(-m^2+2m-d)^2} \underbrace{ - \sqrt{q}  \frac{mn}{1+\Delta} c_2 \frac{(-2m + 2 ) \log(d)}{(-m^2+2m-d)^2}}_{h_d(m,n,\Delta)} \\
& \quad -\frac{u}{2}\big[ c_2 \frac{\log(d)}{-m^2 + 2m - d}  - c_2 \frac{\log(d)}{1 - d} + c_1m \log(\frac{1+\beta}{1-\alpha}) - c_1 \log(1+\beta)\big]_{+}^2\\
& \quad +\underbrace{\frac{u}{2}\big[ c_2 \frac{\log(d)}{-m^2 + 2m - d}  - c_2 \frac{\log(d)}{1 - d} + c_1m \log(\frac{1+\beta}{1-\alpha}) - c_1 \log(1+\beta)\big]_{+}^2}_{g_d(m,n)^2} = 0,
\end{align*}
where for notational simplicity $m = m(t)$, $n = n(t)$, $\Delta = \Delta(t)$ and $u = u(t)$.
\end{proof}
\end{lemma}
More specifically, note that $\lim_{d \rightarrow 1} l_d(m(t), n(t), p(t)) = \frac{1}{2}(1 - m(t))^2 u(t) + \frac{1}{2} q\,p^2(t) u(t)$
and thus 
\begin{equation*}
p^*(t) = \lim_{d \rightarrow 1} p^*_d(t) = \frac{1}{\sqrt{q}} \left[ c_1 m(t) \log{\frac{1 + \beta}{1 - \alpha}} - c_1 \log{(1 + \beta)} - c_2)\right]
\end{equation*}
Then, if we set $c_1 = \frac{1}{\log{\frac{1 - \alpha}{1 + \beta}}}$ and $c_2 = \frac{\log{(1 - \alpha)}}{\log{\frac{1 - \alpha}{1 + \beta}}}$,
we can readily conclude that the optimal selection probability is given by:
\begin{align}
p^*(t) = \frac{1}{\sqrt{q}} (1 - m(t))
\label{eqn:p-opt}
\end{align}
Finally, note that $1 - m(t) \in [0, 1]$ and hence, as long as $q \ge 1$, we have $p^*(t) \le 1$, which will satisfy the constraints required in 
Eq.~\ref{eq:optimal-p}.

\section{Additional details on iTheorie F\"{u}hrerschein Auto} \label{app:itheorie}
Learners use the iTheorie F\"{u}hrerschein Auto to prepare for the written section of the driving lessons.
When a learner installs the app, she is assigned to one of the three item selection algorithms randomly via Google Analytics.
The learner does not know which item selection algorithm has been used to create his or her study sessions.

\begin{figure*}[t]
  \centering
  \subfloat[Lesson selection]{\includegraphics[width=0.235\textwidth]{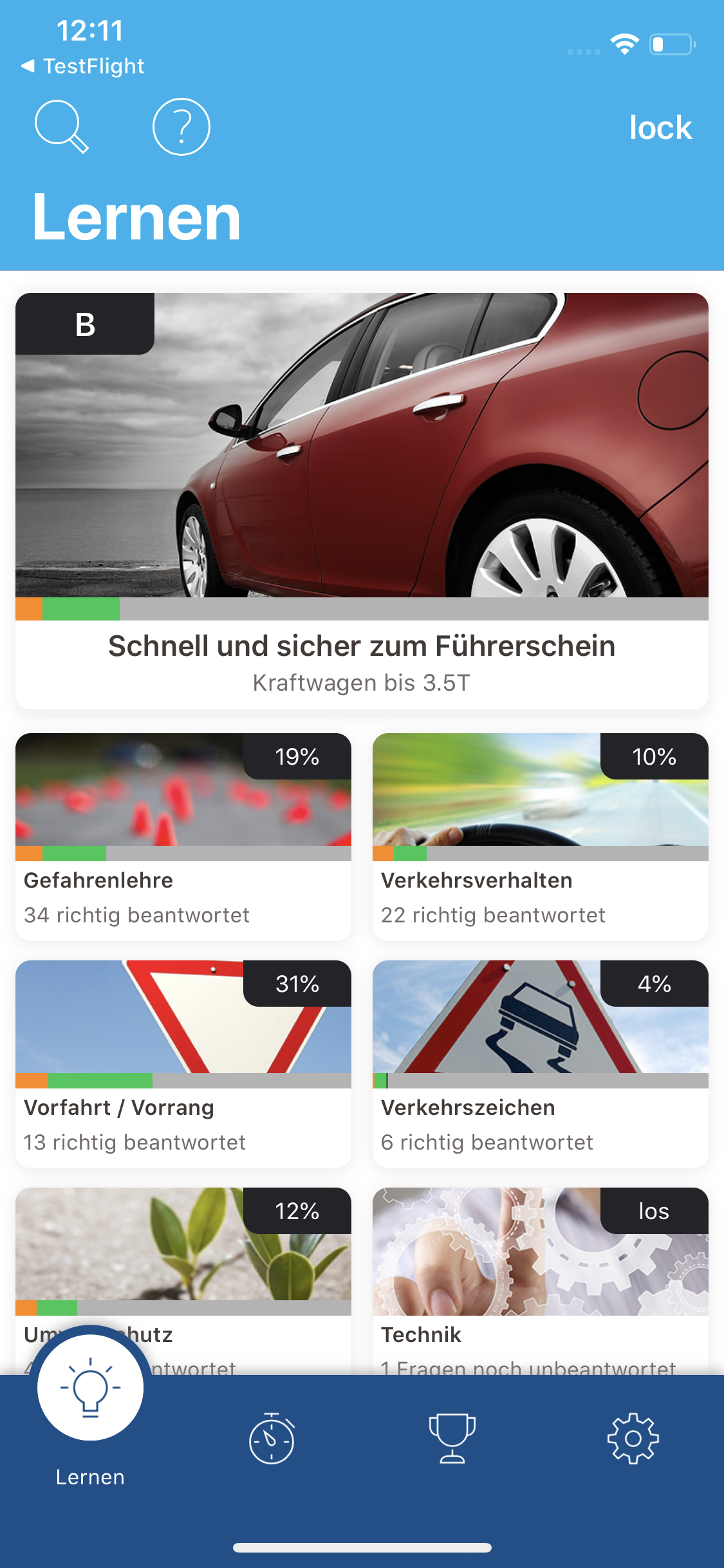}\label{fig:skill-tree}}\hspace{0.01\textwidth}%
  \subfloat[Session start]{\includegraphics[width=0.235\textwidth]{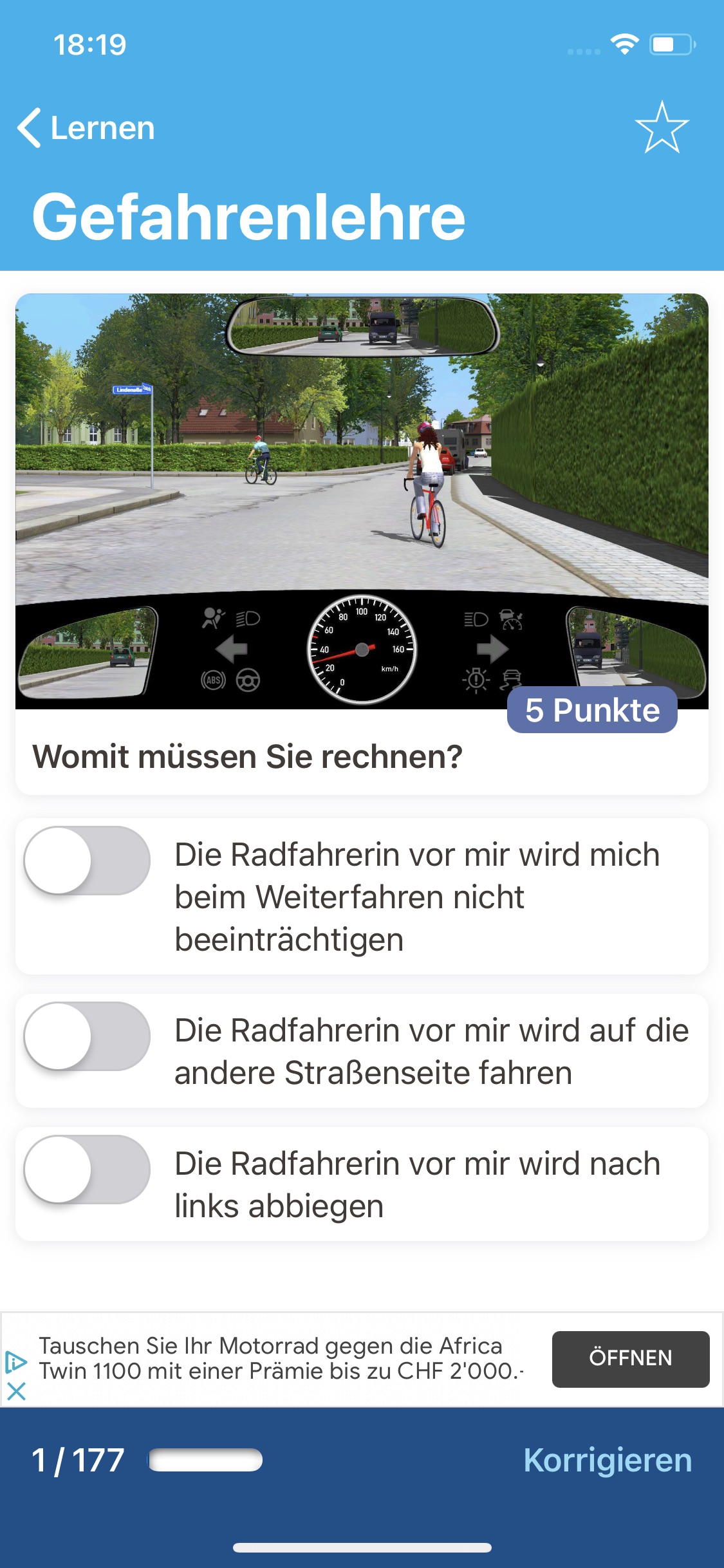}\label{fig:session-start}}\hspace{0.01\textwidth}%
  \subfloat[Correct response]{\includegraphics[width=0.235\textwidth]{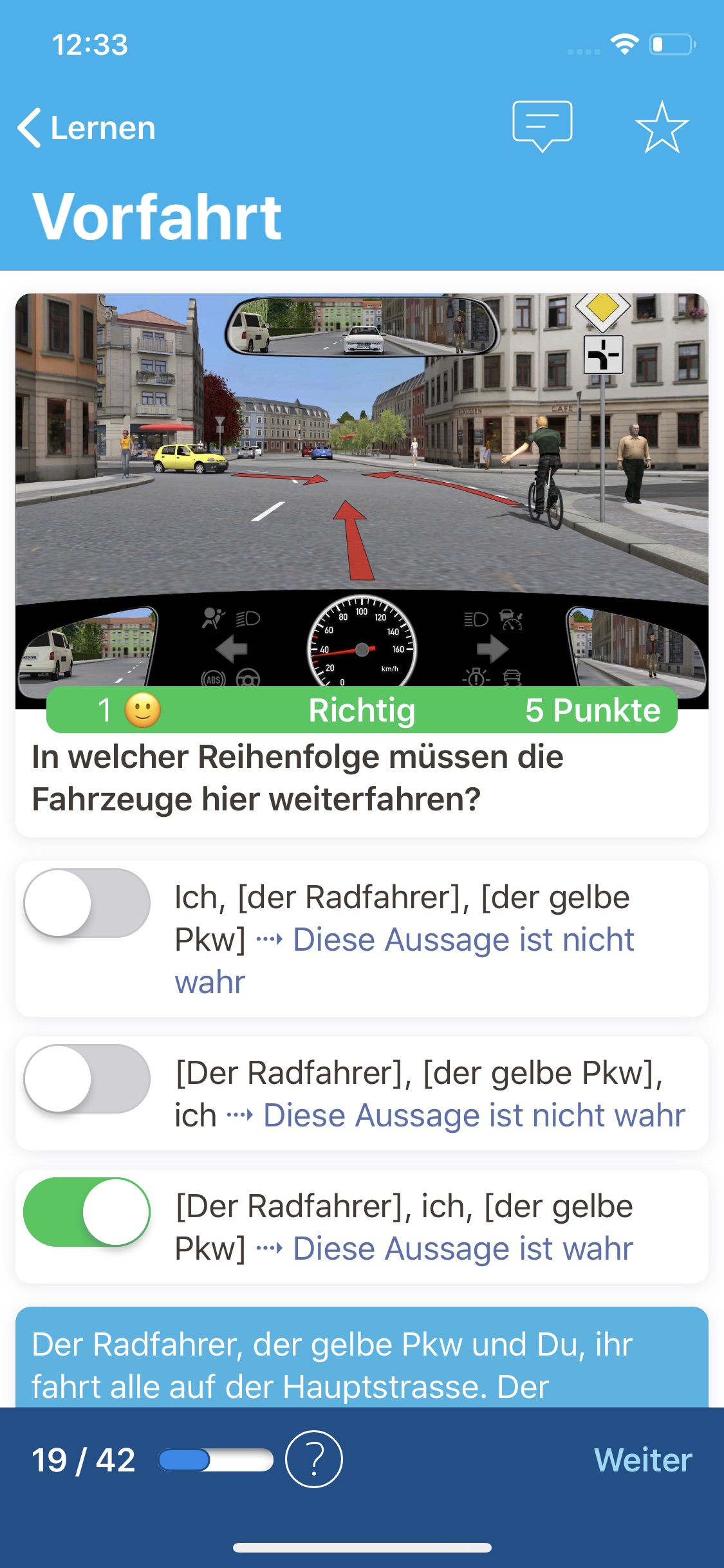}\label{fig:correct-response}}\hspace{0.01\textwidth}%
  \subfloat[Incorrect response]{\includegraphics[width=0.235\textwidth]{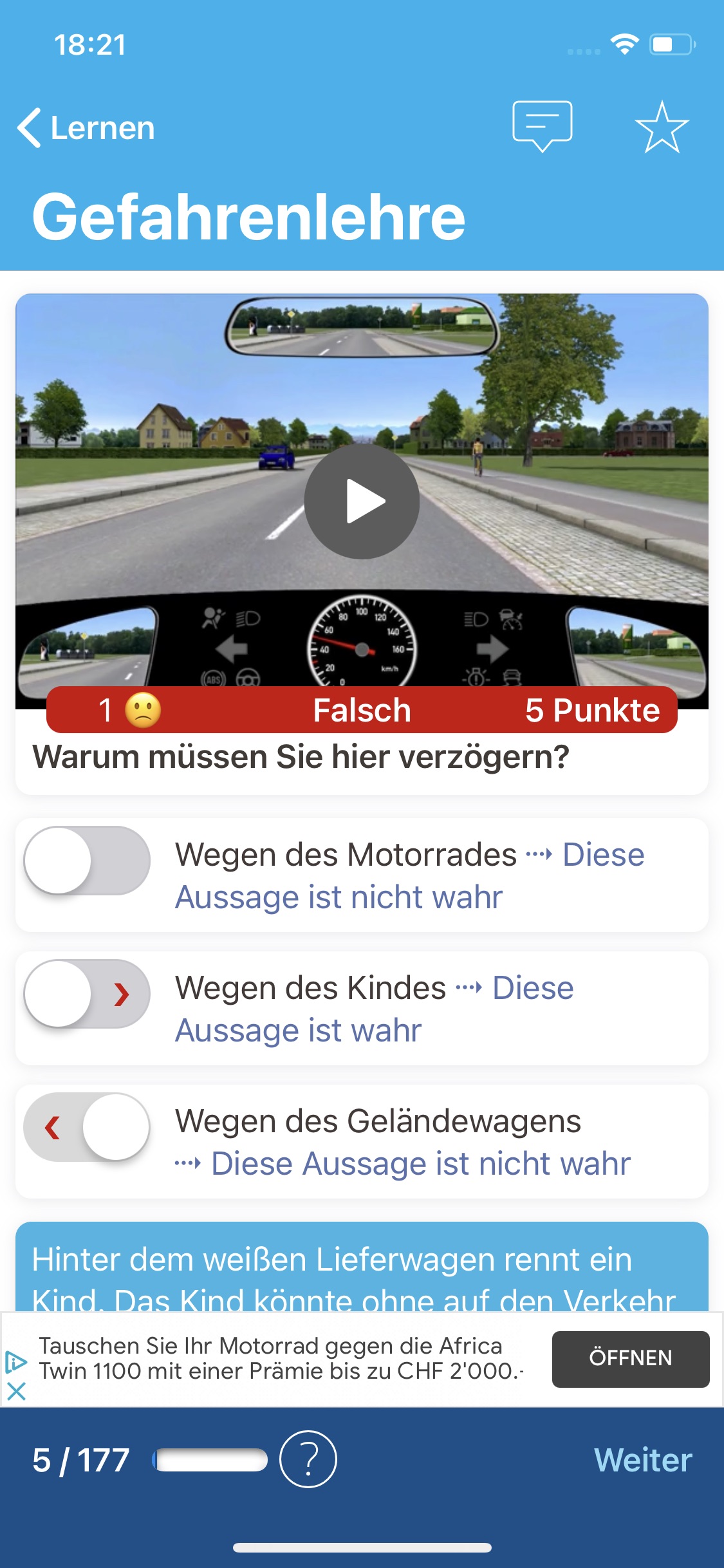}\label{fig:incorrect-response}}
  \caption{The iTheory learning app.}\label{fig:itheory-session}
\end{figure*}

Upon starting the app, the learners are greeted with a screen where they can select the lessons they would like to take, shown in Figure~\ref{fig:skill-tree}.
Once they select a category, a study session starts and they are given questions to answer, as shown in Figure~\ref{fig:session-start}.
The selection of items in each section is done using the algorithm assigned to the user by Google Analytics.
A study session continues until the learner takes a break of longer than 5 minutes.
Figures~\ref{fig:correct-response} and~\ref{fig:incorrect-response} show the notification shown to the user after a correct and incorrect answer respectively.

\section{Predictive performance of the memory model} \label{app:prediction}
% \begin{itemize}
% \item Reiterate what the memory model is.
% \item Describe how we learn the parameters briefly.
% \item Talk about the hyper-parameter optimization.
% \item Show the results.
% \end{itemize}

In this section, we evaluate the predictive performance of the exponential and the power-law memory models using data from February 2019 to June 2019. In both cases, we fit the model 
parameters $n_i(0)$, $\alpha$ and $\beta$ using the variant of half-life regression proposed by Tabibian \ea. [See Appendix, Section 8] and performed a grid search to determine the optimal 
values of the hyper-parameters.

Table~\ref{tab:empirical-evaluation} summarizes the results. In terms of mean average error (MAE) in predicting the recall of items and in terms of correlation between the predicted and empirically 
observed half-life for the items (COR${}_h$), the exponential model clearly outperforms the power-law model. However, in terms of Area Under the Curve (AUC), the power-law model performs slightly 
better than the exponential model. 

Given these results, we decided to use the exponential memory model to estimate the recall probability during our randomized controlled trial.
%
% Curiously, this was in contrast with the Power law memory model performing the best for the Duolingo dataset with Tabibian \ea.
%
\begin{table}[h]
   \centering
   \begin{tabular}{r|c|c}
      % & \multicolumn{1}{c}{\textbf{HLR}} & \multicolumn{1}{|c}{\textbf{Our Model}} & \multicolumn{1}{|c}{\textbf{Our Model}} \\
      & \multicolumn{1}{c}{Exponential} & \multicolumn{1}{|c}{Power-law} \\\hline
      MAE$\downarrow$ & \textbf{0.139} & 0.282 \\
      AUC$\uparrow$ & 0.887 & \textbf{0.901}  \\
      COR${}_h\uparrow$ & \textbf{0.611} & 0.571  \\
   \end{tabular}
   \caption{Predictive performance of the exponential and power-law forgetting curve models. The arrows indicate whether a higher
   value of the metric is better ($\uparrow$) or a lower value ($\downarrow$).}
   \label{tab:empirical-evaluation}
\end{table}

% Our memory model can be expressed in the following form:
% %
% \begin{align}
% m_i(t) = \PP(r_i(t)) &= (1 + \omega (t - t_r) )^{-n_i(t)} & \text{Power law}\\
% m_i(t) = \PP(r_i(t)) &= \exp( -n_i(t) \times (t - t_r) ) & \text{Exponential}
% \end{align}

% We use the variant of the HLR used by Tabibian \ea.  as to learn the parameters of our memory model.
%
%
% We assume Exponential memory model and

\end{document}